# Causal Relationship Network of Risk Factors Impacting Workday Loss in Underground Coal Mines

Shangsi Ren [1], Cameron A. Beeche [1], Zhiyi Shi [1], Maria Acevedo Garcia [2], Katherine Zychowski [3], Shuguang Leng [4], Pedram Roghanchi [2], Jiantao Pu[1, 5*].

[1]Department of Radiology, University of Pittsburgh, Pittsburgh, PA 15213, USA

[2]Department of Mineral Engineering, New Mexico Institute of Mining and Technology
Socorro, NM 87801, USA

[3]College of Nursing, University of New Mexico, Albuquerque, NM 87106, USA

[4]Department of Internal Medicine, University of New Mexico, Albuquerque, NM 87106, USA

[5]Department of Bioengineering, University of Pittsburgh, Pittsburgh, PA 15213, USA

**\*Corresponding authors and guarantors of the entire manuscript:**

Jiantao Pu, PhD
3240 Craft Place
Pittsburgh, PA 15213
jip13@pitt.edu
(412) 641-2571

**Acknowledgments:**

This study was supported in part by grants from the CDC-NIOSH (#75D3012C12182), (#75D30119C06390), and the National Institutes of Health (NIH) (R01CA237277, U01CA271888, R61AT012282).

**Abstract**

**Purpose:** To establish the causal relationship network between various factors leading to workday loss in underground coal mines using a novel causal artificial intelligence (AI) method.

**Methods:** The analysis utilizes data obtained from the National Institute for Occupational Safety and Health (NIOSH). A total of 101,010 injury records from 3,982 unique underground coal mines spanning the years from 1990 to 2020 were extracted from the NIOSH database. An expert review was employed to select relevant variables for analysis. Causal relationships were analyzed and visualized using a novel causal AI method called Grouped Greedy Equivalence Search (GGES). The impact of each variable on workday loss was assessed through intervention do-calculus adjustment (IDA) scores. Model training and validation were performed using the 10-fold cross-validation technique. Performance metrics, including adjacency precision (AP), adjacency recall (AR), arrowhead precision (AHP), and arrowhead recall (AHR), were utilized to evaluate the models.

**Results:** Findings revealed that after 2006, key direct causes of workday loss among mining employees included *total mining experience*, *mean office employees*, *mean underground employees*, *county*, and *total mining experience* (*years*). *Total mining experience* emerged as the most influential factor, whereas *mean employees per mine* exhibited the least influence. The analyses emphasized the significant role of *total mining experience* in determining *workday loss*. The models achieved optimal performance, with AP, AR, AHP, and AHR values measuring 0.694, 0.653, 0.386, and 0.345, respectively.

**Conclusion:** This study demonstrates the feasibility of utilizing the new GGES method to clarify the causal factors behind the workday loss by analyzing employment demographics and injury records and establish their causal relationship network. By leveraging the underlying causal relationship network, appropriate staffing decisions can be made to reduce the risks of workday loss in coal mines.

I. **Introduction**

Coal mine safety is a universally recognized global concern [1-3] with underground mining being considered a high-risk occupation in the United States [3-5]. Despite the overall improvement in safety [6], the mining industry continues to be associated with a high degree of uncertainty. Variations in operational risk within the industry stem from factors such as mine size, workforce size, coal production, and investments in monitoring and safety systems [7, 8].

Many risk assessment models have been developed to improve employee safety and identify potential hazards in the mining industry [9, 10]. Paltrinieri et al. [11] have outlined five essential requirements for an "ideal" risk assessment model, while Beeche et al. [12] have introduced a classification model that utilizes machine learning to assess and categorize coal mines based on mining personnel injuries. Li et al. [13] used Bayesian networks to evaluate the risk of an underground gas explosion, presenting coal mining explosions as a quantitative measure of coal mine risk. A crucial aspect of risk assessment lies in its capability to adapt and improve dynamically over time, enabling the prediction and response to emerging risks that may not have been previously observed or reported.

The National Institute for Occupational Safety and Health (NIOSH) at the Centers for Disease Control and Prevention (CDC) diligently maintains comprehensive records encompassing injury counts, mining site details, employment statistics, production figures, and work hours for every mine operating in the United States. These valuable data are stored in two distinct databases: Address/Job database and Accident/Injury/Illness database [14]. While the risks associated with mining primarily arise from natural threats, they can manifest in various forms. In this study, we propose to use a novel causal artificial intelligence (AI) method to establish the causal relationship network of the risk factors affecting the duration of work absence among miners due to job-related injuries [15].

Workday loss has been recognized by Margolis [16] as an indicator of mine risk, representing the duration during which miners are absent from work due to work-related injuries. Margolis conducted an analysis of the Mining Safety and Health Administration (MSHA) data spanning from 2003 to 2007, revealing a statistical significant association between age, total mining experience, and workday loss [16]. Coleman and Kerkering [17], in their work, advocated for utilizing damage-statistic variables as indices for mining safety, with workday loss serving as an appropriate index. Through the analysis of MSHA data from 1983 to 2004 and employing the beta distribution to simulate losses, they concluded that workday loss holds considerable importance as a relevant index [17]. Furthermore, Nowrouzi-Kia et al. [18] conducted a comprehensive analysis of nine mining industry articles pertaining to workday loss, identifying various antecedents that can impact workday loss. These antecedents include the mining work environment, gender, age, mining equipment, organizational size, disease status, job training, recovery time, pre-injury health, and access to medical care. Their findings highlighted the relationship between different types of workday loss and specific mining types

In this study, we propose to utilize an innovative causal AI algorithm to analyze the potential causes contributing to the duration of work-related injury absences among miners. To conduct this analysis, we leveraged the data from the NIOSH Address/Employment and Accident/Injury/Illness databases, aiming to establish a causal relationship network among these causes using a directed graph approach. We consider the mine as a fully connected system [19], incorporating personal characteristics of employees (work experience, mining work experience, working age, mining work age), working characteristics of the mine (underground work area, above-ground work area, public work District, overall coal mine output, overall number of employees), mine-specific employee attributes (employee title, employee disease type, employee injury type) as variables of equal importance. The evaluation of causal discovery results was performed using "pattern metrics" [15], including adjacency precision (AP), adjacency recall (AR), arrowhead precision (AHP), and arrowhead recall (AHR). The primary focus of this study is to explore and identify potential causes contributing to employee sickness/injury absences and establish their corresponding causal relationship network. By doing so, we aim to offer decision-makers valuable technical guidance for coal mine risk assessment and subsequent risk management.

## II. Materials and Methods

### A. Study datasets

To perform this study, we used data from the NIOSH address/employment and accident/injury/illness records to analyze mining operations data since 1983. The dataset comprised various variables extracted from coal mine records, including: (1) mine ID, (2) county, (3) workday loss, (4) underground employees, (5) surface at underground mine employees, (6) mill or prep plant employees, (7) office employees, (8) total number of employee hours, (9) total tons of clean coal produced, (10) accident injury illness, (11) accident type, (12) total mine experience, (13) total experience in one specific mine, (14) regular job experience, (15) job title, (16) source of injury, (17) degree of injury, (18) shift time hours. We focused on NIOSH records for coal mining operations spanning from 1990 to 2020 (inclusive), excluding earlier years due to concerns regarding inconsistent reporting and significant technological advancement that greatly improved working conditions. A total of 101,010 data entries were analyzed, each corresponding to annual employment information reported to NIOSH for mining operations. Mines with missing data or those that were contractor-operated were excluded from the study. During the data preprocessing stage (Fig. 1), experts of coal mining field assisted in encoding non-numeric variables into categorical variables (Table 1). Following the data selection, the distribution of the data was presented in Table 2.

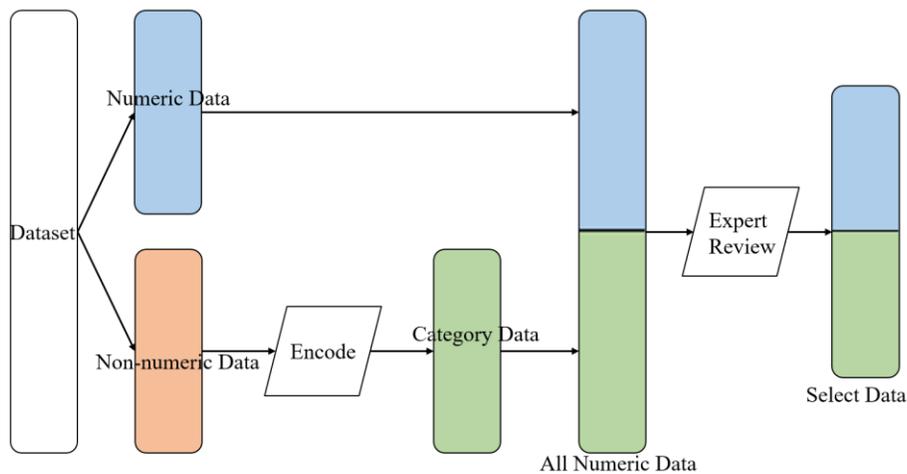

**FIGURE 1.** Data preprocess and variables select process.

**Table 1.** Encoding of six non-numeric variables from the NIOSH coal mine address/employment and accident/injury/illness reports from 1990-2020 (n = 101,010).

| Variables | Original Non-numeric Type | Encode Type* |
|---|---|---|
| county | 217 | 217 |
| accident injury illness | 47 | 32 |
| accident type | 75 | 44 |
| job title | 121 | 9 |
| source of injury | 127 | 127 |
| degree of injury | 11 | 11 |

*Encode details are in Supplemental Table 1 - Supplemental Table 6.

**Table 2.** Distribution of 21 numeric variables from the NIOSH coal mine address/employment and accident/injury/illness report from 1990-2020 (n=101,010).

| Variables | Mean | Variance | SD* | Range |
|---|---|---|---|---|
| underground employees | 208.231 | 3.004E+04 | 173.331 | 0, 1142 |
| surface employees | 15.522 | 278.767 | 16.696 | 0, 259 |
| mill or prep plant employees | 15.018 | 650.163 | 25.498 | 0, 164 |
| office employees | 7.181 | 93.736 | 9.682 | 0, 145 |
| underground work hours | 4.630E+05 | 1.527E+11 | 3.908E+05 | 0, 2.817E+06 |
| surface work hours | 3.476E+04 | 1.299E+09 | 3.604E+04 | 0, 3.042E+05 |
| mill or prep plant work hours | 3.323E+04 | 3.206E+09 | 5.662E+04 | 0, 3.358E+05 |
| office work hours | 1.597E+04 | 4.982E+08 | 2.232E+04 | 0, 3.410E+05 |
| total working hours | 5.445E+05 | 2.180E+11 | 4.669E+05 | 0, 3.016E+06 |
| total tons of coal | 1.870E6 | 4.142E+12 | 2.035E+06 | 0, 1.446E+07 |
| total mine experience | 14.795 | 78.624 | 8.867 | 0.192, 92.057 |
| total experience in one specific mine | 7.299 | 58.783 | 7.667 | 0.192, 90.575 |
| regular job experience | 6.461 | 45.593 | 6.752 | 0, 95.613 |
| shift time hours | 4.442 | 9.871 | 3.142 | 0, 23.983 |
| County | 93.667 | 4024.636 | 63.44 | 100, 9998 |
| accident injury illness | 12.277 | 22.018 | 4.692 | 1, 32 |
| accident type | 15.491 | 125.658 | 11.21 | 0, 43 |
| *job title* | 2.912 | 6.506 | 2.551 | -1, 8 |
| *source of injury* | 80.207 | 1021.58 | 31.962 | 1, 127 |
| *degree of injury* | 2.110 | 1.541 | 1.241 | 0, 10 |
| *workday loss* | 33.425 | 4132.250 | 64.283 | 0, 7600 |

*SD stand for Standard Deviation

### B. Causal relationship network modeling based on grouped greedy equivalence search (GGES)

Causal discovery or network modeling is a process that utilizes data to uncover the underlying causal relationship between variables within a dataset [20]. Traditional methods for causal discovery have strict requirements for the dataset and variables [21], demanding a large sample size and complete information about the variables. These requirements pose challenges when applying these methods to real-world datasets, often leading to inaccurate conclusions or illogical causal relationships. To overcome these limitations, we introduced a novel method called Grouped Greedy Equivalence Search (GGES). GGES leverages prior knowledge to enhance the identification of local optimal solution within a dataset.

GGES, a score-based method, stems from the Greedy Equivalence Search (GES) Method[22]. The GGES method requires the dataset to satisfy the Markov condition [23] and the faithfulness hypothesis [24]. In the GGES method, the score of the graphical model is calculated using the Bayesian Information Criterion (BIC) score [25] and a generalized score [26]. The implementation of the GGES method has three steps (Fig. 3):

**(2)** All the variables are divided into two groups; one group for causal variables, the other for predict variables. The predicted variables are assumed to not influence any other variables within their group and only accept influence from other variables. The causal variables can produce and receive effects on each other. This grouping process requires some prior knowledge about which variables represent the outcomes.

$$V = v_1, v_2, \dots, v_n \supseteq Vc + Vp \tag{1}$$

$$C \begin{cases} v_i \rightarrow v_j, & v_i \in Vc, v_j \in Vp \\ v_j \nrightarrow v_i, & v_i \in Vc, v_j \in Vp \\ v_i \rightarrow v_j, & (v_i, v_j) \in Vc \\ v_i \nrightarrow v_j, & (v_i, v_j) \in Vp \end{cases} \quad (2)$$

Where $V$ is the variables, $v_n$ is one variable in $V$, $Vc$ is the causal variables group, $Vp$ is the predict variable group, $C$ is the constraint condition.

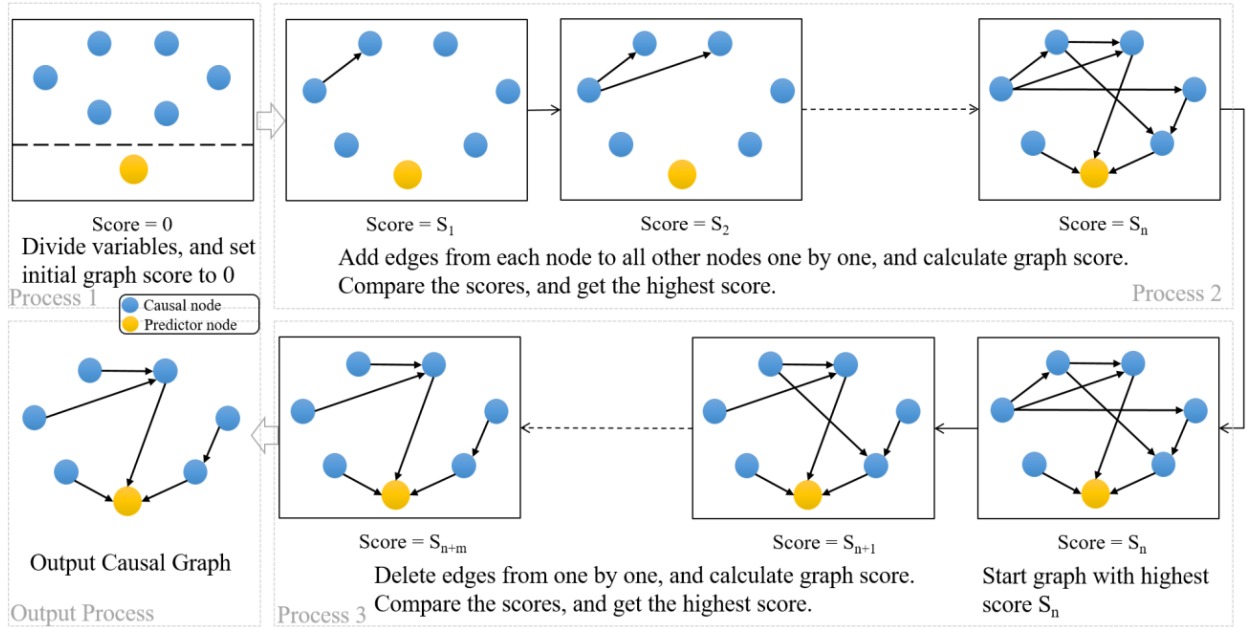

**FIGURE 2.** Overview of the GGES method. **Step 1:** variables are divided into two groups: causal variables and predict group, and an initial graph score of 0 is set. **Step 2:** Edges are added and new graph score is calculated, with the graph with the higher score being kept. **Step 3:** Edges are removed, and the new graph score is calculated, with the graph with higher score being kept. The final causal relationship network is obtained after these steps.

(2) The GGES method starts with an initial empty graph and sets the score of the whole graph to 0. Edges are added from nodes sequentially, and the graph score is calculated and compared to select the model with the highest score. Nodes assigned to the outcome variable group are skipped according to the constraint placed at the beginning of the method.

(3) From the graph model with the highest score, each edge is deleted in turn, and a new graph model is calculated. The graph model with the highest graph score is selected as the final causal relationship network result.

$$\max f(G, D) \ s.t. \ G \in \Omega, G| = C \tag{3}$$

Where $f$ is the structure scoring function, $\Omega$ is the structure space, and $G| = C$ means that $G$ satisfies the constraint condition $C$. In the process of searching and scoring, the constraint condition $C$ is to require the searched structure to satisfy the acyclic structure in the structure graph.

### C. Training GGES models

The GGES method was used to construct a causal relationship network, elucidating the connections between multiple independent variables and the actual workday loss experienced by mining employees. Our approach involved a comprehensive analysis of the causal relationships between all variables and workday loss time. Subsequently, we conducted individual assessments of the variables directly associated with the workday loss relationship, aiming to discern the causality of the variables that exerted the most significant influence at each stage.

To evaluate the contributions between variables, we utilized the intervention do-calculus adjustment (IDA) score [27] as the foundation for our judgments. By integrating the independent analyses of each component with an overarching analysis, our final model provided a comprehensive and accurate representation of the most suitable causal model. This approach enabled us to gauge the relative impact and interplay of the variables, facilitating a robust understanding of the factors influencing workday loss among mining employees.

### D. Performance validation

Models were trained and tested on annual data across two distinct time intervals: (1) a span of 31 years from 1990 to 2020, and (2) a period of 15 years from 2006 to 2020. To ensure robustness, all models underwent validation using the 10-fold cross-validation technique. During the training phase, the causal discovery model was fed with the training dataset to generate a causal relationship network. The test dataset was then utilized to assess the causal outcomes, employing evaluation metrics such as adjacency precision (AP), adjacency recall (AR), arrowhead precision (AHP), and arrowhead recall (AHR) [15]. Additionally, the Total Causal Effects (TCE) value [28] was calculated for the variables included in the model to describe the magnitude of the relationship between the variables. Higher absolute TCE values indicate stronger associations between the variables. All statistics were performed in R 3.4.1 and Python. A p-value less than 0.05 was considered statistically significant.

### III. Results

#### A. Variable Selection

The study dataset consisted of 101,010 data entries from 3,982 mines from 1990 to 2020 with 64,597, 22,009, and 14,404 from 1990-1999, 2000-2009, and 2010-2020, respectively (Table 3). Our final data cohort included 119 variables from 2,917 miners. After removing non-significant variables, 21 variables were identified (Table 2). Among the 21 variables, 15 original data type variables were analyzed for data distribution (Table 3).

**Table 3.** Results of statistical analysis of NIOSH coal mine address/employment and accident/injury/illness reports from 1990-2020. P-value corresponds to the Pearson correlation coefficient between the variable's total column and # of days off.

| *Variable* | *1990-1999 (n=64,597)* | *2000-2009 (n=22,009)* | *2010-2020 (n=14,404)* | *Total (n=101,010)* | *p-value\** |
|---|---|---|---|---|---|
| mean underground employees | 195.73 (75.84) | 202.66 (91.29) | 272.82 (124.16) | 208.23 (93.24) | **<0.05** |
| mean surface employees | 16.71 (8.03) | 13.07 (7.41) | 13.93 (9.65) | 15.52 (8.26) | 0.630 |
| mean mill/plant employees | 13.98 (9.29) | 15.15 (10.54) | 19.45 (13.19) | 15.02 (10.61) | **<0.05** |
| mean office employees | 6.93 (4.50) | 7.10 (4.89) | 8.44 (6.72) | 7.18 (5.17) | **<0.05** |
| mean employees per mine | 52.97 (90.81) | 237.98 (107.48) | 314.65 (143.09) | 245.95 (109.41) | **<0.05** |
| total hours worked | 499,951 (199,867) | 551,920.27 (253,199) | 733,069 (337,675) | 544,517.23 (252,455) | **<0.05** |
| underground tons mined | 1,524,169 (788,360) | 2,225,817.13 (1277,428) | 2,875,101.99 (1619,574) | 1,869,693 (1162,488) | **<0.05** |
| experience years in this job title | 6.68 (6.93) | 6.51 (6.86) | 5.41 (5.76) | 6.46 (6.75) | **<0.05** |
| experience years at this mine | 8.50 (8.77) | 5.67 (5.86) | 4.42 (4.72) | 7.32 (7.67) | **<0.05** |
| total mining experience (years) | 15.58 (9.22) | 14.67 (8.64) | 11.49 (6.02) | 14.80 (8.88) | **<0.05** |
| shift time hours | 4.42 (3.11) | 4.46 (3.18) | 4.51 (3.22) | 4.44 (3.13) | **<0.05** |
| regular job experience | 658.01 (686.52) | 641.92 (681.44) | 531.37 (574.62) | 636.26 (678.25) | **<0.05** |
| experience at this mine | 836.42 (864.55) | 553.29 (584.63) | 429.06 (485.33) | 718.94 (767.73) | **<0.05** |
| total mining experience | 1547.38 (919.47) | 1458.35 (885.64) | 1140.04 (553.92) | 1470.71 (889.50) | **<0.05** |
| workday loss | 29.73 (15.78) | 37.48 (10.13) | 43.81 (8.79) | 33.43 (13.09) | **<0.05** |

[1]**Mean±(SD); n (%)**
**\*p-value represents the Pearson correlation coefficient between each total variable and the total workday loss.**

### B. Causal analysis of workday loss using all the variables from 1990-2020

Utilizing all available variables, we employed the GGES method to construct a comprehensive causal relationship network. After removing variables that were unrelated to *workday loss*, we obtained a clear causal relationship network of *workday loss* with 17 variables (Fig. 3). Among these variables, the following were identified as direct causes of workday loss: (1) *mean underground employees*, (2) *mean employees per mine*, (3) *mean office employees*, (4) *source of injury*, (5) *total mining experience*, (6) *accident type*, (7) *total mining experience* (*years*), (8) *mean surface employees*, (9) *job title*, and (10) *county*.

Notably, as demonstrated in Table 4, *total mining experience* emerged as the most influential cause, while *mean employees per mine* exhibited the least influential impact on workday loss.

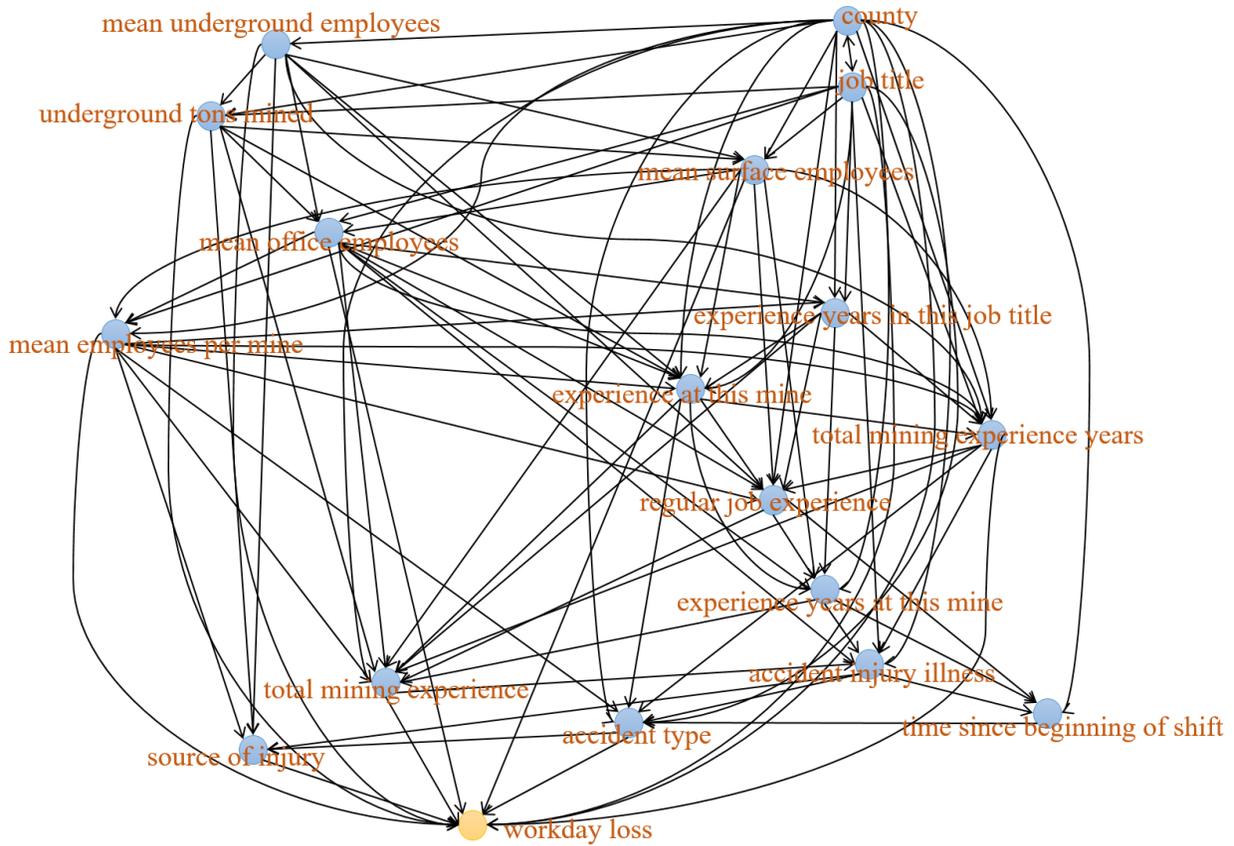

**FIGURE 3.** Directed causal graph or network between 17 variables with *workday loss* from 1990-2020.

**TABLE 4.** TCE value and Pearson Correlation between 17 variables with *workday loss*. Bolded rows indicate statistical significance ($p < 0.05$).

| *Variables* | *Correlation* | *p-value* | *TCE Value* |
|---|---|---|---|
| *mean employees per mine* | **0.042** | **<0.05** | **-0.193** |
| *job title* | **-0.034** | **<0.05** | **-0.293** |
| *experience at this mine* | **0.032** | **<0.05** | **-0.629** |
| *experience years in this job title* | **0.028** | **<0.05** | **-0.629** |
| *mean surface employees* | 0.002 | 0.630 | -1.872 |
| *experience years at this mine* | **0.032** | **<0.05** | **-1.872** |
| county | **-0.016** | **<0.05** | **-2.307** |
| total mining experience | **0.069** | **<0.05** | **-2.354** |
| underground tons mined | **0.174** | **<0.05** | **3.327** |
| source of injury | **0.174** | **<0.05** | **3.327** |
| mean office employees | **-0.012** | **<0.05** | **-4.112** |
| accident type | **0.086** | **<0.05** | **-4.445** |
| mean underground employees | **0.046** | **<0.05** | **5.345** |
| time since beginning of shift | **-0.010** | **<0.05** | **5.347** |
| total mining experience (years) | **0.069** | **<0.05** | **8.473** |
| accident injury illness | **0.023** | **<0.05** | **8.998** |
| total mining experience | **0.032** | **<0.05** | **192.9027** |

Based on the TCE value in Table 2 and the evaluation procedure, the AP, AR, AHP, and AHR were found to be 0.646, 0.628, 0.368, and 0.354, respectively.

## C. Causal analysis of workday loss using all the variables from 1990-2005

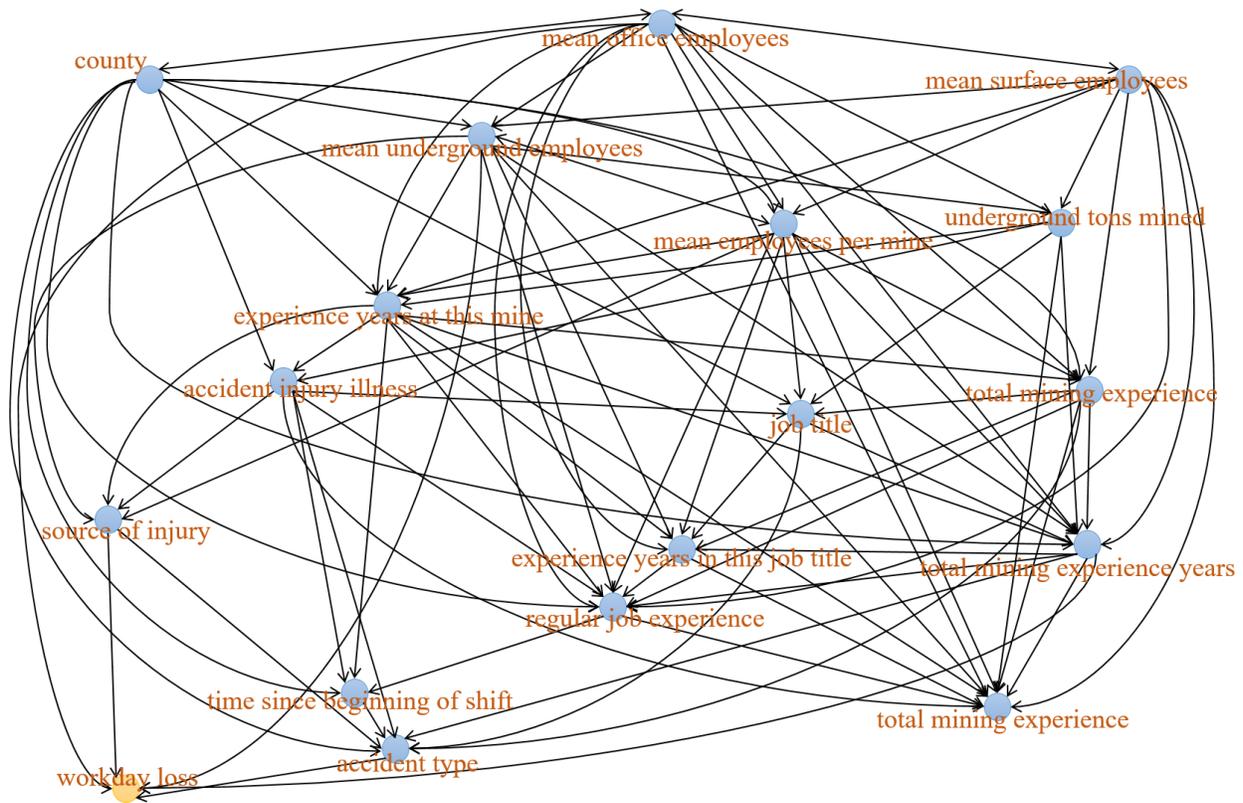

**FIGURE 4.** Causal relationship network between 17 variables with *workday loss* from 1990-2005.

The causal relationship network from 1990-2005 (Fig. 4) is highly consistent with the causal relationship network from 1990-2020. The causal relationship network from 1990-2005 illustrates that (1) *mean office employees*, (2) *mean underground employees*, (3) *source of injury*, (4) *accident type*, and (5) *total mining experience (years)* are the direct cause of *workday loss*.

**D. Causal analysis of workday loss using all the variables from 2006-2020**

The direct cause of *workday loss* include: (1) *total mining experience*, (2) *mean office employees*, (3) *mean underground employees*, (4) *county*, and (5) *total mining experience (years)* (Fig. 5). According to Table 5, *total mining experience* is the most influential cause, while *mean employees per mine* is the least influential cause.

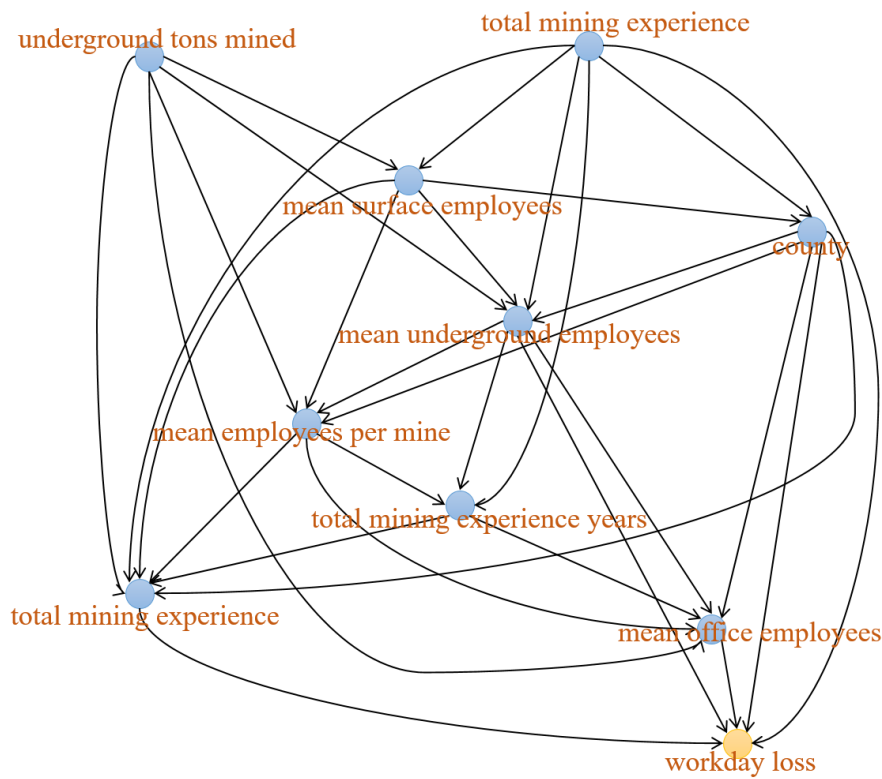

**FIGURE 5.** Directed causal relationship network between 9 variables with *workday loss* from 2006-2020

**TABLE 5.** TCE value and Pearson Correlation between 9 variables with *workday loss* from 2006-2020. Bolded rows indicate statistical significance ($p < 0.05$).

| variables | correlation | p value | TCE Value |
|---|---|---|---|
| *mean employees per mine* | **0.042** | **<0.05** | **-0.19384** |
| *experience at this mine* | 0.032 | <0.05 | -0.6294 |
| *mean surface employees* | 0.002 | 0.630 | -1.87229 |
| *county* | **-0.016** | **<0.05** | **-2.30391** |
| *underground tons mined* | **0.174** | **<0.05** | **3.327258** |
| *mean office employees* | -0.012 | <0.05 | -4.11196 |
| *mean underground employees* | 0.046 | <0.05 | 5.345713 |
| *total mining experience (years)* | 0.069 | <0.05 | 8.473438 |
| *total mining experience* | 0.032 | <0.05 | 192.9027 |

Based on the TCE value in Table 2 and the evaluation procedure, the AP, AR, AHP, and AHR were 0.694, 0.653, 0.386, and 0.345, respectively.

### IV. Discussion

We developed a novel causal discovery algorithm termed "Grouped Greedy Equivalence Search" (GGES) and applied it to analyze coal mine data across the United States using the NOISH database. To account for the impact of coal mine safety implemented in 2006 [29], we divided the data into pre-2006 and post-2006. Using this approach, a causal relationship network for employee workday loss was established to elucidate the factors influencing employee workday loss in coal mines and their relationship. This novel methodology goes beyond traditional approaches and provides a valuable and insightful contribution to the understanding of workday loss dynamics in mining operations, offering new perspectives for risk assessment and management strategies in the industry.

Our results revealed several key factors that directly impact workday loss among mining employees. The variables found to have a direct influence on workday loss include: (1) *total mining experience*, (2) *mean office employees*, (3) *mean underground employees*, (4) *county*, and (5) *total mining experience*

*(years)*. Among these factors, *total mining experience* emerged as the most significant contributor to *workday loss.* This suggests that as employees gain more experience in the mining industry, they are more likely to experience an absence due to illness or injury. However, this finding could also suggest that as employees gain experience, they opt for more dangerous and risk-prone tasks in underground mines. Time since the start of a shift was found to have an inverse correlation with workday loss due to illness or injury (table 4). This suggests that the most dangerous injuries are likely to occur when shifts are changing in mining operations. This highlights the importance of considering employees' personal experiences when allocating staff and implementing safety measures in coal mines. Additionally, our results also reveal that the distribution of personnel in a mine can also contribute to the workday loss, with a higher number of employees in the underground mining area and fewer employees in the surface and office areas being associated with a higher risk of employee's absences due to illness or injury. This suggests that while technical advancements have improved mine safety, there still remains a significant risk in the underground mining area.

We employed the GGES method to uncover the causal relationships between various features in complex mines. These features are often correlated, making it challenging to establish connections using traditional statistical analysis methods. Of the 22 variables used in this study, six were non-numeric variables and were encoded as categorical data to be included in our models. Traditional statistical correlation heavily relies on data integrity [30]. The GGES method, however, can identify local optima and explore possible causal relationships even with incomplete data. The use of 10-fold cross-validation method in our study resulted in high evaluation scores, with AP and AR values above 0.6, indicating consistency between our causal map results and the data. This further reflects the accuracy of our findings.

In comparison with Margolis's study [16], our research expands upon traditional statistical methods by incorporating additional variables and exploring their causal relationship. While our findings are in alignment with Margolis' results [16], our study goes beyond by considering a broader range of factors that contribute to workday loss among mining employees. Unlike the focus of Nowrouzi-Kia et al.'s study [17], which incorporated personal characteristics such as age and gender, our research specifically delves into variables related to mine distribution. However, our results are consistent with those of Nowrouzi-Kia et al. [17] in highlighting the impact of variables such as mining type, management scale, and job training on workday loss. This concurrence strengthens the validity and reliability of our findings.

There are limitations associated with the study. First, we did not take into account other potential factors that might affect *workday loss* by miners. For instance, the impact of mining equipment and methods, such as the longwall method or hand method, was not considered due to the difficulty in obtaining and differentiating this information. Second, the data collected in this study spanned over 31 years, during which time changes in regulation and safety practices may have affected the accuracy and completeness of the data [31]. Additionally, several mines fail to report either (1) employment or (2) accident information causing incomplete employee injury reports to be discarded. This could suggest that mines without accurate data reporting are at increased risk. However, we believe that these limitations do not significantly impact the important findings of this study. Third, for categorical variables not represented by data, we relied on expert opinions to reduce the number of categories, but this may still impact the results. Despite these limitations, we believe our study has effectively captured the causes of workday loss among miners, including individual employee characteristics, mine job characteristics, and employee mine characteristics.

## V.     Conclusion

Our study aims to understand the causal factors that contribute to the actual workday loss of mining employees based on data from the National Institute for Occupational Safety and Health (NIOSH). To do this, we developed a novel algorithm called Grouped Greedy Equivalence Search (GGES) to analyze and visualize the causal relationships among variables in the dataset. A total of 101,010 entries from 31 years of reports were used in our analyses. Our results indicate that (1) total mining experience, (2) mean office employees, (3) mean underground employees, (4) county, and (5) total mining experience (years) are the direct causes of workday loss, with total mining experience being the most influential factor. We also found that the distribution of personnel in the mine, specifically the number of personnel in the underground mining area, can have an impact on workday loss. By understanding the key factors contributing to workday loss, mine operators can make informed decisions on staffing and safety measures to minimize risks and reduce the number of workday losses due to injury or illness. Our study provides a comprehensive and detailed understanding of causality in the coal mining industry, which can be used to improve the safety and efficiency of coal mines in the future.